\documentclass[11pt]{article}

\usepackage[preprint]{acl}

\usepackage{times}
\usepackage{latexsym}

\usepackage[T1]{fontenc}
\usepackage[utf8]{inputenc}

\usepackage{microtype}

\usepackage{inconsolata}

\usepackage{graphicx}

\usepackage{amsmath}
\usepackage{amsthm}
\usepackage{amssymb}
\usepackage{algorithm}
\usepackage{algorithmic}
\usepackage{etoolbox}
\usepackage{multirow}
\usepackage{booktabs}
\usepackage{bm}

\usepackage{mathtools} % provides \splitfrac, \mathclap, etc.
\usepackage{microtype} % already included above; kept for line-breaking quality
\usepackage{adjustbox} % for \begin{adjustbox}{max width=\linewidth}
\usepackage{tabularx}  % flexible-width columns
\usepackage{makecell}  % easier line breaks inside table cells
\usepackage{ragged2e}  % \RaggedRight for tabularx columns

\allowdisplaybreaks
\title{Reachability-Aware Pretraining for Efficient Target-Oriented Path Exploration in Temporal Knowledge Graph Reasoning}

\author{
\textbf{Chien-Liang Liu\textsuperscript{1}} \and \textbf{Tsao-Lun Chen\textsuperscript{2,3}}
\\
\\
 \textsuperscript{1}Artificial Intelligence Research Center, Chang Gung University,
 \\
 \textsuperscript{2}Graduate Institute of Oral Biology, National Taiwan University,
 \\
 \textsuperscript{3}Electrical Engineering, National Taiwan University of Science and Technology
\\
   \texttt{liucl@cgu.edu.tw}, \texttt{r14450024@ntu.edu.tw}
}

\begin{document}
\maketitle
\begin{abstract}

Temporal Knowledge Graph (TKG) reasoning under the extrapolation setting focuses on forecasting future time-stamped events (facts) from historical data in a temporal knowledge graph. Existing approaches, reinforcement learning (RL)-based multi-hop reasoning methods are prominent for TKG reasoning because they produce human-interpretable predictions via explicit multi-hop path tracing. However, during RL training, rewards are typically sparse, and exploration is highly inefficient due to the vast, time-evolving action space. These issues hinder efficient training and often limit overall performance. To address these challenges, we propose \textbf{RAPTOR} (\textbf{R}eachability-\textbf{A}ware \textbf{P}retraining for Efficient \textbf{T}arget-\textbf{OR}iented Path Exploration), a self-supervised pretraining method that injects a reachability-aware inductive bias to the agent. By learning to estimate the reachability of candidate actions to the target entity, RAPTOR reduces exploration over unpromising paths and provides a strong initialization for downstream RL fine-tuning.
Experimental results on the ICEWS14, ICEWS05-15, and ICEWS18 datasets demonstrate that RAPTOR pretraining markedly improves the training efficiency and consistently outperforms conventional baselines, establishing it as an effective approach for enhancing RL-based multi-hop reasoning methods for TKG reasoning.

\end{abstract}

\section{Introduction}

\label{sec1}

\begin{figure}[!t]
    \centering
    \includegraphics[width=1\linewidth]{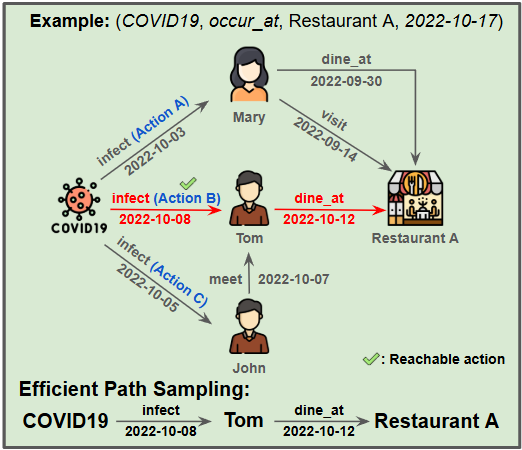}
    \caption{Example of multi-hop reasoning in TKGs.}
    \label{fig1}
\end{figure}

Temporal Knowledge Graphs (TKGs) have emerged as a pivotal graph-based data structure for modeling dynamic relationships that evolve over time. Formally, a TKG is composed of events, and each event is represented in a quadruple (subject, relation, object, timestamp). Compared to static knowledge graphs, TKGs incorporate the temporal information, enabling systems to capture how events change over time. As a result, TKGs have been widely used in downstream applications, such as recommendation systems \cite{TKGrecommendation} and question answering \cite{saxena2021cronkgqa,ijcai2023p571}.

In this work, we focus on TKG reasoning under the extrapolation setting, where the goal is to answer a query $(e_q,r_q,?,t_q)$ by predicting the missing entity using only events with timestamps $t<t_q$. Prior work has proposed a variety of approaches for this task \cite{CyGNet,jin-etal-2020-recurrent,RE_GCN}; however, these methods are typically less interpretable because they do not explicitly expose the underlying reasoning process. In contrast, multi-hop reasoning methods \cite{sun-etal-2021-timetraveler,Tpath,Dream,TAgent} output explicit temporal paths as evidence, which can reveal \emph{why} a prediction is made and provide human-checkable supporting events. As shown in Figure~\ref{fig1}, given a query $(\textit{COVID19}, \textit{occur\_at}, ?, \text{2022-10-17})$, a multi-hop reasoner predicts the missing object by outputting a path that starts from the subject entity ("\textit{COVID19}") and ends at the predicted entity (e.g., "\textit{Restaurant A}").

However, multi-hop reasoning methods comes with a major optimization challenge. At each time step, the agent must choose from a large set of timestamped outgoing edges, and the number of possible trajectories grows exponentially with the hop budget. Moreover, TKG reasoning usually incorporates temporal validity (e.g., non-decreasing timestamps) so the action space is not only large but also \emph{time-dependent}: an outgoing edge at one entity may become invalid due to timestamp constraints. As a result, most sampled trajectories terminate without receiving a reward, leading to sparse terminal rewards and slow, unstable reinforcement learning (RL) training.

To address this, we propose \textbf{RAPTOR}, to the best of our knowledge the first self-supervised pretraining method tailored for RL-based multi-hop reasoning on TKGs. RAPTOR injects a \emph{reachability-aware} inductive bias into the agent by pretraining the policy to estimate whether each candidate action can reach the target entity within the hop budget while satisfying temporal constraints. This pretraining objective directly mitigates the wasted exploration problem: unreachable actions (i.e., those with no temporally valid path to the target entity within $K$ steps) are down-weighted early, so the subsequent RL phase focuses on feasible parts of the graph.
% 看要不要加
% Since RAPTOR is implemented as a pretraining stage, it can be combined with existing techniques without modifying their training objectives, providing a general plug-in initialization for diverse TKG reasoners.

Our contributions: (i) we introduce a reachability-based self-supervised task for temporal multi-hop reasoning, (ii) we design an efficient labeling procedure to construct reachability supervision under temporal constraints, and (iii) we demonstrate consistent improvements in both convergence speed and final accuracy on ICEWS14, ICEWS05-15, and ICEWS18.

\section{Related works}

\subsection{Multi-hop Reasoning on Static Knowledge Graphs}

Multi-hop reasoning on static knowledge graphs is often trained with RL, but it typically suffers from sparse terminal rewards and large action spaces. To improve exploration, prior work introduces denser guidance before or during RL. DeepPath \cite{xiong-etal-2017-deeppath} mines paths between source and target entities using bidirectional breadth-first search (BFS) and then pretrains the agent via supervised imitation of these paths, which increases the probability of sampling rewarding trajectories during RL. Multi-hop KG \cite{lin-etal-2018-multi_hop_KG} performs reward shaping with embedding-based scores, assigning soft rewards to non-target entities to provide more informative feedback. SSRL \cite{KGR_SL} uses BFS-derived label to identify actions with high-expected-return at each state and pretrains the policy to encourage efficient exploration. However, these methods do not consider temporal constraints and thus cannot be applied to TKG reasoning.

\subsection{Temporal Knowledge Graph Reasoning}

Unlike static knowledge graphs, TKGs attach timestamps to events, which leads to different settings (interpolation and extrapolation) in TKG reasoning. The interpolation setting aims to predict missing object entity at historical records \cite{TTtransE,TA-DistMult,Lacroix2020Tensor,Goel_Kazemi_Brubaker_Poupart_2020}. In contrast, the extrapolation setting focuses on forecasting future events using past observations; in this work, we adopt the extrapolation setting. Previous works explore diverse modeling paradigms for extrapolation. For example, \citet{han2021explainable} studies explainable TKG reasoning by providing human-interpretable rationales, while \citet{jin-etal-2020-recurrent,RE_GCN} model temporal dynamics with recurrent and relational GNN architectures. CyGNet \cite{CyGNet} further adopts a generative forecasting framework to predict future events from historical sequences.

\subsection{Multi-hop Reasoning on Temporal Knowledge Graphs}

% Similar to KG reasoning, multi-hop reasoning on Temporal Knowledge Graphs (TKGs) is typically trained with reinforcement learning (RL), but it must additionally satisfy temporal constraints. As a result, the agent operates in a time-dependent action space, which makes RL optimization challenging under sparse rewards. As a basic RL formulation, TAgent \cite{TAgent} applies the multi-hop agent framework to TKGs without using additional auxiliary signals. To improve training efficiency, TITer \cite{sun-etal-2021-timetraveler} incorporates time-shaped rewards to guide the agent toward temporally consistent reasoning behaviors. TPath \cite{Tpath} adds a path reward that encourages repeatable and efficient path patterns, thereby narrowing the search space during training. DREAM \cite{Dream} mines demonstration paths with bidirectional BFS and leverages adversarial imitation learning to help the agent discover feasible reasoning paths more efficiently. However, these techniques introduce additional guidance \emph{during} RL training, meaning the agent must learn effective exploration behaviors on the fly, which can slow training and increase the reliance on careful signal design.

Similar to knowledge graph reasoning, multi-hop reasoning on TKGs is typically trained with RL, but it must additionally satisfy temporal constraints. As a result, the agent operates in a time-dependent action space, which makes RL optimization challenging under sparse rewards. As a basic RL formulation, TAgent \cite{TAgent} applies the multi-hop agent framework to TKGs without using additional auxiliary signals. Prior work has also attempted to improve training efficiency; for example, TITer \cite{sun-etal-2021-timetraveler} incorporates time-shaped rewards, TPath \cite{Tpath} adds a path reward, and DREAM \cite{Dream} leverages demonstration paths with adversarial imitation learning. However, these techniques introduce additional guidance \emph{during} RL training, meaning the agent must learn effective exploration behaviors on the fly, which can increase the reliance on careful signal design due to the potential interference with other training objectives.

\section{Preliminaries} \label{sec2}

\subsection{Temporal Knowledge Graphs}
\label{DF_TKG}

A TKG is a collection of time-stamped events, denoted by graph $\mathcal{G} = (\mathcal{E}, \mathcal{R}, \mathcal{T}, \mathcal{Q})$, where $\mathcal{E}$, $\mathcal{R}$, $\mathcal{T}$, and $\mathcal{Q}$ represent the sets of entities, relations, timestamps, and events, respectively. The set of events $\mathcal{Q} \subseteq \mathcal{E} \times \mathcal{R} \times \mathcal{E} \times \mathcal{T}$ contains quadruples $(e_s, r, e_o, t)$, representing a relation $r$ between entity $e_s$ and entity $e_o$ at timestamp $t$. In this work, the notation $t < \hat{t}$ denotes that timestamp $t$ precedes $\hat{t}$ (i.e., the event at $t$ occurs earlier than the one at $\hat{t}$).

Following \citet{Tlogic,sun-etal-2021-timetraveler}, we augment the TKG with inverse edges. For each quadruple $(e_s, r, e_o, t) \in \mathcal{Q}$, we add a corresponding inverse quadruple $(e_o, r^{-1}, e_s, t)$ to the graph, where $r^{-1}$ denotes the inverse relation of $r$.

\subsection{Temporal Path}

%可以參考TPAR A Unified Temporal Knowledge Graph Reasoning Model Towards Interpolation and Extrapolation的寫法，只是改成有constraint

% Following the settings in \cite{sun-etal-2021-timetraveler,han2021explainable}, we define a temporal path $P$ of length $l \in \mathbb{N}$ as a sequence of $l$ quadruples connecting a source entity to a target entity. A path is considered \textbf{temporally valid} only if the timestamps of consecutive edges follow a non-decreasing order. Formally, a temporal path $P$ is denoted as:
% Following the settings in \cite{sun-etal-2021-timetraveler,han2021explainable},
We define a temporal path as a sequence of quadruples that is temporally valid. Following the settings in \citet{sun-etal-2021-timetraveler,han2021explainable,lin-etal-2023-techs}, a temporal path is considered valid only if the timestamps of consecutive quadruples follow a non-decreasing order. Formally, a temporal path of length $n$ is denoted as a sequence of quadruples:
\begin{equation}
    \begin{gathered}
    \bigl((e_{i-1}, r_i, e_i, t_i)\bigr)_{i=1}^{n}, \\
    \text{where } t_{i-1} \le t_i,\ \forall i \in \{1, \dots, n\}.
    \end{gathered}
\end{equation}
where $t_0$ is the initial timestamp, $e_0$ represents the source entity, and $(e_{i-1}, r_i, e_i, t_i) \in \mathcal{Q}$ represents the $i$-th quadruple in the path.

\subsection{Problem Formulation}

Given a graph $\mathcal{G}$ and a query $q = (e_q, r_q, ?, t_q)$, the goal of TKG reasoning is to infer the missing object entity $o_{q}$. As mentioned in Section~\ref{sec1}, we focus on the extrapolation setting. In this setting, for a query at timestamp $t_q$, the model is restricted to accessing historical events that occurred before the query timestamp. Formally, the set of available historical events $\mathcal{H}$ is defined as:

\begin{equation}
    \mathcal{H} = \{ (e_s, r, e_o, t) \in \mathcal{Q} \mid t < t_q \}
\end{equation}

Accordingly, we train the model to estimate the conditional probability of candidate entities and maximize the likelihood of the ground-truth entity $o_q$ given $q$ and $\mathcal{H}$:
\begin{equation}
    \max_{\theta} P(o_{q} \mid e_q, r_q, t_q, \mathcal{H})
\end{equation}

Specifically, RL-based multi-hop reasoning methods typically formulate the reasoning task as a Markov Decision Process (MDP). We cast the prediction process as a sequential decision-making task, where the agent starts at $e_q$ and reaches the target $o_q$ after executing consecutive actions, as illustrated in Figure~\ref{fig1}.

% For instance, consider a query $(\textit{COVID-19}, \textit{Occur}, ?, \text{2022-10-17})$. By discovering the following temporal path, the model predicts the quadruple $(\textit{COVID-19}, \textit{Occur}, \textit{Restaurant A}, \text{2022-10-17})$:

% \begin{equation}
%     \begin{aligned}
%         \textit{COVID-19} \xrightarrow[\text{2022-10-08}]{\textit{Infect}} \textit{Tom}
%         \xrightarrow[\text{2022-10-12}]{\textit{Dine at}} \textit{Restaurant A}
%     \end{aligned}
% \end{equation}

\subsection{Reinforcement Learning Framework}

In RL phase, we formulate the prediction process as a MDP. The MDP process is represented by a 4-tuple $(S, A, P, R)$, where $S$, $A$, $P$ and $R$ denote the state space, action space, transition function, and reward function, respectively. Each component is elaborated as follows.

\subsubsection{States}

The state $s \in S$ is defined as a 5-tuple $(\xi,k,e_k,t_k,v_k)$. Here, $\xi = (e_q, r_q, t_q)$ is the query information. The scalar $k \in \{0,\ldots,K\}$ denotes the time step in the MDP and also the current path length, where $K$ is the maximum path length. The pair $(e_k, t_k)$ denotes the current entity and timestamp. Finally, $v_k = \{e_0,\ldots,e_k\}$ is the set of visited entities. At initialization, we set $e_0=e_q$, $t_0=0$, and $v_0=\{e_0\}$. We reindex timestamps by shifting the earliest time to 0, so $t_0=0$ corresponds to the earliest timestamp in the graph. Therefore, the initial state is $(\xi,0,e_0,t_0,v_0)$.

\subsubsection{Actions}

The action space $A$ is defined as a set of all possible actions. The set of candidate actions $A_s \subseteq A$ is dependent on the current state $s$. In practice, $A_s$ corresponds to outgoing edges from $e_{k}$. To encourage efficiency and non-redundant exploration, we enforce the path to be a simple path by masking actions that revisit previously visited entities, except the self-loop action \textsc{Stop}. Specifically, we define $A_s = \{(e', r', t') \mid(e_{k}, r', e', t') \in Q, t_k \le t' < t_q,e' \not\in v_k\} \cup \{\textsc{Stop}\}$. The self-loop action \textsc{Stop} $=(r_{stop}, e_{k}, t_{k})$ allow agent to stay in place.

% An episode terminates when the agent selects \textsc{Stop} or reaches the maximum length k.

\subsubsection{Transition}

The environment updates the state according to the action selected by the agent. The transition function $P: S \times A \rightarrow S$ is defined as follows. Given a state $s$ and an action $a=(e', r', t') \in A_s$, the next state $s'$ is
\[
s' = P(s, a) = (\xi, k+1, e_{k+1}, t_{k+1}, v_{k+1}),
\]
where $e_{k+1}=e'$, $t_{k+1}=t'$, and the visited-entity set is updated as $v_{k+1}=v_k \cup \{e'\}$.
If $k=K$, the episode terminates without taking the $K$-th action.

\subsubsection{Rewards}

The agent receives a terminal reward $R = 1$ if the episode terminates at the ground-truth answer entity $o_{q}$, and $0$ otherwise:
\begin{equation}
R =
\begin{cases}
1, & \text{if episode terminates at } o_q,\\
0, & \text{otherwise},
\end{cases}
\end{equation}
We train the policy and value networks using the actor-critic algorithm; full optimization details are provided in Appendix~\ref{sec:appendix_actor_critic}.

\begin{figure*}[!h]
    \centering
    \includegraphics[width=1\linewidth]{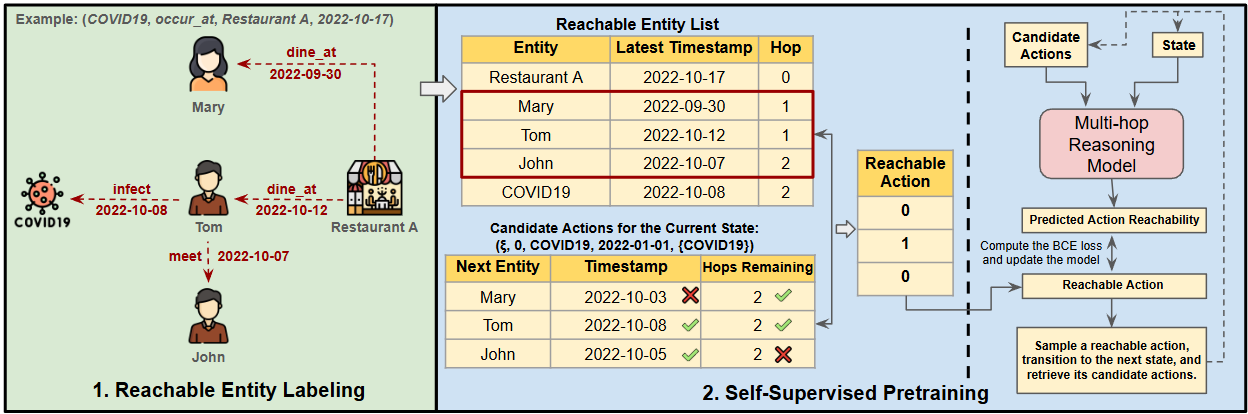}
    \caption{Overview of RAPTOR, where the maximum path length is set to 2 for illustration.}
    \label{fig2}
\end{figure*}

\section{Methodology}

We propose RAPTOR, a reachability-aware pretraining framework for temporal multi-hop reasoning. The overall pretraining pipeline (Figure~\ref{fig2}) consists of two stages: (i) constructing reachability supervision under temporal constraints via a reachable entity labeling algorithm, (ii) pretraining the policy to identify the candidate actions that can reach the target entity $o_q$ within the hop budget. After pretraining, we fine-tune the pretrained agent with RL. In this section, we describe the policy network, the reachable-entity labeling algorithm, and the self-supervised pretraining objective.

\subsection{Policy Network}

We use a simple policy network with an LSTM path encoder and MLP heads to evaluate RAPTOR. We concatenate the path, query, and step embeddings, then pass the combined vector through a two-layer MLP to form a state representation. This state vector is used to score each candidate action. The detailed architecture is described as follows.

Given a state $s = (\xi, k, e_k, t_k, v_k)$ and the candidate set $A_s$, the actor outputs a probability distribution $\pi_\theta(\cdot\mid s)$ over the candidate actions in $A_s$, and the critic estimates $V_\phi(s)$.

\paragraph{Embeddings.}

We represent each entity $e$ with a static embedding $\mathrm{Emb}_\mathcal{E}(e)\in\mathbb{R}^{d_e}$. For relations, we use time-conditioned embeddings. Following \citet{sun-etal-2021-timetraveler}, we define a temporal encoding
\begin{equation}
\boldsymbol{\tau}(\Delta t)=\cos(\mathbf{w}\Delta t+\mathbf{b}),
\end{equation}
where $\mathbf{w},\mathbf{b}\in\mathbb{R}^{d_t}$ are learnable parameters and $\Delta t = t_q-t$.

Given a relation $r$ with timestamp $t$ and query time $t_q$, we map $r$ to its static embedding $\mathrm{Emb}_\mathcal{R}(r)$ and form a time-conditioned relation embedding by concatenation:
\begin{equation}
\mathbf{r}(r,\Delta t)=[\mathrm{Emb}_\mathcal{R}(r);\boldsymbol{\tau}(\Delta t)]\in\mathbb{R}^{d_r}.
\end{equation}

For action $a=(e',r',t')$, we construct the action embedding as
\begin{equation}
\label{eq14}
\mathbf{a}(a) = \big[\mathbf{e'};\mathbf{r}(r',t_q-t')\big]\in\mathbb{R}^{d_a},
\end{equation}
where $d_a = d_e + d_r$. We also introduce a static step embedding $\mathrm{Emb}_{\text{step}}(k) \in \mathbb{R}^{d_{\text{step}}}$ to encode the time step and provide the information of remaining hop budget.

\paragraph{Temporal path encoding.}
% \begin{equation}
% \mathbf{x}_i=\big[\ \mathbf{e}_{i}\ ;\ \mathbf{r}(r_i,\Delta t_i)\ \big]\in\mathbb{R}^{d_e+d_r}, \quad i=0,\ldots,\ell.
% \end{equation}
We encode the temporal path into an embedding using a single-layer LSTM. At time step $k$, we have an ordered sequence of taken actions $\{a_0,a_1,\ldots,a_{k-1}\}$. For each action $a_i$ in this sequence, we construct its action embedding $\mathbf{a}_i = \mathbf{a}(a_i)$ as described in Eq.~\eqref{eq14}.

When $k=0$, there is no executed action yet, so we introduce a learnable initial action input, denoted $\mathbf{a_{\mathrm{init}}}$. We then define the input sequence as $\{a_{\mathrm{init}}, a_0,\ldots,a_{k-1}\}$ and use the LSTM to update the temporal path embedding as follows:
\begin{equation}
\begin{aligned}
\mathbf{h}_0
&= \operatorname{LSTM}(\mathbf{a}_{\mathrm{init}}, \mathbf{h}_{\mathrm{init}}), \\
\mathbf{h}_{i+1}
&= \operatorname{LSTM}(\mathbf{a}_i, \mathbf{h}_i), \forall i \in \{0, \ldots, k - 1\}.
\end{aligned}
\end{equation}
where $\mathbf{h}_{\mathrm{init}} \in \mathbb{R}^{d_a}$ is initialized as a learnable vector.
% Here, $\mathbf{h}_{0} \in \mathbb{R}^{d_e+d_r}$ is a learnable vector.

\paragraph{State vector.}
To construct the state vector, we first concatenate the embeddings of the query entity, query relation, temporal path, and time step to form
\begin{equation}
\mathbf{u}
=
\Big[
\mathrm{Emb}_\mathcal{E}(e_s);
\mathbf{r}(r_q,0);
\mathbf{h}_{k};
\mathrm{Emb}_{\text{step}}(k)
\Big].
\end{equation}
We then project $\mathbf{u}$ to $\mathbb{R}^{d_a}$ with a two-layer MLP to obtain $\mathbf{s}$:
\begin{equation}
\begin{aligned}
\mathbf{s}
&= \mathrm{MLP}(\mathbf{u}) \\
&= \mathbf{W}_{\text{head}}\,\sigma(\mathbf{W}_{\text{hidden}}\mathbf{u}+\mathbf{b}_{\text{hidden}})+\mathbf{b}_{\text{head}},
\end{aligned}
\end{equation}
where $\sigma(\cdot)$ denotes $\mathrm{ReLU}$, and $\mathbf{s}\in\mathbb{R}^{d_a}$ matches the dimensionality of the action embeddings.

\paragraph{Action scoring.}
At time step $k$, we score each candidate action by measuring the similarity between the state vector $\mathbf{s}$ and the embedding of the corresponding action. Specifically, for each $a\in A_s$, we define its logit as
\begin{equation}
z(a)=\langle \mathbf{s},\mathbf{a}(a)\rangle.
\end{equation}
We then obtain the action distribution by applying a softmax over the logits of all candidate actions:
\begin{equation}
\pi_{\theta}(a\mid s)=\frac{\exp\left(z(a)\right)}{\sum_{a'\in A_s}\exp\left(z(a')\right)}.
\end{equation}

\paragraph{Value estimation.}
In addition, we learn a critic to estimate the expected return from the current state. Given $\mathbf{u}$ at time step $k$, the critic applies a two-layer MLP to produce a prediction of scalar value:
\begin{equation}
V_{\phi}(\mathbf{u})=\mathrm{MLP}_{\mathrm{crt}}(\mathbf{u})\in\mathbb{R},
\end{equation}
which serves as an estimate of the expected cumulative return used for advantage computation during training.

\subsection{Reachable Entity Labeling}

To label candidate actions during self-supervised pretraining, we perform \emph{reachable entity labeling} for all $(o_q,t_q)$ pairs in the training set. For each pair, we run a backward BFS from $o_q$ to find entities that can reach $o_q$ within the hop limit while respecting temporal constraints. The procedure is summarized in Algorithm~\ref{alg:reachable_labeling}.

Given a graph $\mathcal{G}$, a training example $(e_q, r_q, o_q, t_q)$, max path length $K$, and max in-edges $N$, we initialize a queue with $(o_q, t_q, v, 0)$, where $v$ is the visited set (initialized as $\{o_q\}$) and the hop count is $0$. We then repeatedly dequeue a tuple $(e_\mathrm{cur}, t_\mathrm{cur}, v_\mathrm{cur}, \ell)$ and examine its incoming edges. For each incoming edge $(e', r', e_\mathrm{cur}, t')$, we discard it if $e'$ has been visited or if $t' > t_\mathrm{cur}$ which violates temporal constraints. To reduce computation, we keep only the top-$N$ incoming edges with the largest timestamps using \text{SelectTopN}. If $e'$ passes these checks, it is reachable: we update the visited set, enqueue $(e', t', v_{new}, \ell+1)$, and record $(\text{entity}, \text{latest timestamp}, \text{hop distance}) = (e', t', \ell+1)$ in $L$. To avoid duplicate records for the same entity at the same hop distance from $o_q$, \text{IsLatest} checks whether $t'$ is the latest timestamp associated with $e'$ at that hop distance, and \text{RemoveEarlier} removes any existing record of $e'$ with an earlier timestamp at that hop distance.

\begin{algorithm}[h]
\caption{Reachable Entity Labeling}
\textbf{Input} training example $(e_q, r_q, o_q, t_q)$, graph $\mathcal{G}=(\mathcal{E}, \mathcal{R}, \mathcal{T}, \mathcal{Q})$, maximum path length $K$, maximum number of incoming edges $N$ \\
\textbf{Output} The reachable entities list $L$ of pair $(o_q, t_q)$

\label{alg:reachable_labeling}
\begin{algorithmic}[1]

\STATE $queue \leftarrow \text{Queue}()$
\STATE $L \leftarrow \emptyset$
\STATE \textsc{Enqueue}($queue$, $(o_q, t_q, \text{v}=\{o_q\}, 0)$)
\WHILE{$queue \text{ is not empty}$}
    \STATE $(e_\mathrm{cur}, t_\mathrm{cur}, v_\mathrm{cur}, \ell) \leftarrow \text{Dequeue}(queue)$
    \IF{$\ell \ge K$}
        \STATE \textbf{continue}
    \ENDIF
    \STATE $\mathcal{N}_\mathrm{in} \leftarrow \{(e', t') \mid (e', r', e_\mathrm{cur}, t') \in \mathcal{Q} \text{, } t' \le t_\mathrm{cur}\}$
    \STATE $\mathcal{N}_\mathrm{pruned} \leftarrow \text{SelectTopN}(\mathcal{N}_\mathrm{in}, N, \text{key}=t')$
    \FORALL{$(e', t') \in \mathcal{N}_\mathrm{pruned}$}
        \IF{$e' \notin v_\mathrm{cur} $}
            \STATE $v_\mathrm{new} \leftarrow v_\mathrm{cur} \cup \{e'\}$
            \STATE \text{Enqueue}($queue$,$(e', t', v_\mathrm{new}, \ell+1)$)
            \IF{$\text{IsLatest}(e', t', \ell, L)$}
                \STATE $L \leftarrow \text{RemoveEarlier}(e', \ell, L)$
                \STATE $L \leftarrow L \cup \{(e', t', \ell+1)\}$
            \ENDIF
        \ENDIF
    \ENDFOR
\ENDWHILE

\RETURN $L$
\end{algorithmic}
\end{algorithm}

\subsection{Self-Supervised Pretraining}

The goal of this self-supervised pretraining is to push the predicted score of reachable actions toward 1 and unreachable actions toward 0. For a training example $(e_q, r_q, o_q, t_q)$, we first construct the reachable entity list $L$ for the pair $(o_q, t_q)$. At time step $k$, each candidate action $a = (e', r', t') \in A_s$ is labeled as reachable if $L$ contains a tuple $(e', \bar{t}, \bar{h})$ such that $t' \le \bar{t}$ and $K - k > \bar{h}$; otherwise, it is labeled as unreachable. This produces a binary label $y(a) \in \{0, 1\}$ for each candidate action.

We then train the policy network to align with $\bm{y}$ using a binary cross-entropy loss. The reachability of each candidate action is estimated by applying a sigmoid function to its logit:
\begin{equation}
% \hat{y}(a) = \text{Sigmoid}\big(z(a)\big), \quad \forall a \in A_s,
\hat{y}(a) = \operatorname{sigmoid}\big(z(a)\big), \quad \forall a \in A_s,
\end{equation}
where $\hat{y}(a)$ is the predicted reachability of action $a$.
The self-supervised loss at time step $k$ is defined as
\begin{equation}
\begin{split}
\mathcal{L}_{\mathrm{SL}}(\theta)
= - \frac{1}{|A_s|} \sum_{a \in A_s} \Big[
y(a) \log \hat{y}(a) \\
\hspace{0.8em}+ (1 - y(a)) \log (1 - \hat{y}(a))
\Big].
\end{split}
\end{equation}

During pretraining, we transition to the next time step $k+1$ by sampling from the output distribution over reachable actions, and repeat this process until reaching the maximum path length.
The overall self-supervised loss for an episode is
\begin{equation}
\mathcal{L}_{\mathrm{SL}}^{\mathrm{ep}}(\theta)
= \mathbb{E}\left[\sum_{k=0}^{K-1} \mathcal{L}_{\mathrm{SL}}^{(k)}(\theta)\right].
\end{equation}

We optimize the policy network by minimizing $\mathcal{L}_{\mathrm{SL}}^{\mathrm{ep}}(\theta)$ with stochastic gradient descent.

\section{Experiments}

\begin{table*}[h]
    \centering
    \setlength{\tabcolsep}{3pt}
    \caption{Experimental results on ICEWS14, ICEWS05-15, and ICEWS18 under the time-aware filtered setting. The best and second-best results are highlighted in \textbf{bold} and \underline{underlined}, respectively. $*$ indicates that RL-RAPTOR achieves statistically significant improvements compared with Pure-RL (\(p\text{-value}<0.05\)).}
    \label{tab:main_results}
    \begin{adjustbox}{max width=\textwidth}
    \begin{tabular}{l|l*{3}{c}|l*{3}{c}|l*{3}{c}}
        \toprule
        \multirow{2}{*}{Model} & \multicolumn{4}{c}{ICEWS14} & \multicolumn{4}{c}{ICEWS05-15} & \multicolumn{4}{c}{ICEWS18} \\
        & MRR & Hits@1 & Hits@3 & Hits@10 & MRR & Hits@1 & Hits@3 & Hits@10 & MRR & Hits@1 & Hits@3 & Hits@10 \\
        % \midrule
        % TTransE & 13.72 & 2.98 & 17.70 & 35.74 & 15.57 & 4.80 & 19.24 & 38.29 & 8.31 & 1.92 & 8.56 & 21.89 \\
        % TA-DistMult & 25.80 & 16.94 & 29.74 & 42.99 & 24.31 & 14.58 & 27.92 & 44.21 & 16.75 & 8.61 & 18.41 & 33.59 \\
        % DE-SimplE & 33.36 & 24.85 & 37.15 & 49.82 & 35.02 & 25.91 & 38.99 & 52.75 & 19.30 & 11.53 & 21.86 & 34.80 \\
        % TNTComplEx & 34.05 & 25.08 & 38.50 & 50.92 & 27.54 & 9.52 & 30.80 & 42.86 & 21.23 & 13.28 & 24.02 & 36.91 \\
        \midrule
        xERTE & 40.02 & 32.06 & 44.63 & 56.17 & 46.62 & 37.84 & 52.31 & 63.92 & 29.31 & 21.03 & 33.51 & 46.48 \\
        TLogic & \underline{42.5} & 33.2 & \underline{47.6} & \textbf{60.3} & 47.0 & 36.2 & 53.1 & \underline{67.4} & 29.6 & 20.4 & 33.6 & \underline{48.1} \\
        CyGNet & 35.05 & 25.73 & 39.01 & 53.55 & 36.81 & 26.61 & 41.63 & 56.22 & 24.93 & 15.90 & 28.28 & 42.61 \\
        RE-NET & 36.93 & 26.83 & 39.51 & 54.78 & 43.32 & 33.43 & 47.77 & 63.06 & 28.81 & 19.05 & 32.44 & 47.51 \\
        RE-GCN & 40.39 & 30.66 & 44.96 & 59.21 & \underline{48.03} & 37.33 & \underline{53.85} & \textbf{68.27} & \textbf{30.58} & 21.01 & \textbf{34.34} & \textbf{48.75} \\
        \midrule
        Tpath & 38.9 & 29.4 & 42.5 & 54.6 & 45.7 & 36.8 & 51.6 & 63.8 & 27.4 & 21.1 & 31.3 & 42.8 \\
        TAgent & 39.1 & 30.7 & 43.2 & 55.2 & 45.1 & 36.3 & 51.0 & 63.4 & 27.2 & 20.1 & 30.7 & 42.3 \\
        TITer & 40.87 & 32.28 & 45.45 & 57.10 & 47.69 & 37.95 & 52.92 & 65.81 & 29.98 & \textbf{22.05} & 33.46 & 44.83 \\
        TITer-150 & 41.16 & 32.99 & 46.17 & 57.21 & 47.68 & \textbf{38.57} & 53.36 & 65.51 & 29.30 & 21.62 & 33.31 & 44.77 \\
        \midrule
        Pure-RL & 42.41 & \underline{33.78} & 47.58 & 59.46 & 47.74 & 38.24 & 53.50 & 66.54 & 29.87 & 21.64 & 33.79 & 46.85 \\
        RL-RAPTOR & \textbf{42.86}* & \textbf{34.37}* & \textbf{47.93}* & \underline{59.49} & \textbf{48.04}* & \underline{38.51}* & \textbf{53.90}* & 66.79* & \underline{30.25}* & \underline{21.92}* & \underline{34.3}* & 47.45* \\
        \bottomrule
    \end{tabular}
    \end{adjustbox}
\end{table*}

\subsection{Experimental Setup}

We describe the datasets, evaluation metrics, and baselines in this section; implementation details are provided in Appendix~\ref{sec:appendix_impl_details}.

\subsubsection{Datasets}

We evaluate RAPTOR on three standard TKG benchmarks: ICEWS14, ICEWS05-15 \cite{ICEWS14s}, and ICEWS18 \cite{jin-etal-2020-recurrent}. All are derived from ICEWS and cover events in 2014, 2005--2015, and 2018; statistics are reported in Appendix~\ref{sec:appendix_dataset_stats} (Table~\ref{tab:dataset_statistics}). Following prior work, we split each dataset into train/validation/test sets by timestamp.

\subsubsection{Evaluation Metrics}
% 評估指標介紹、正向和反向評估、time-aware設定

We use the following two common metrics in TKG reasoning to evaluate performance:
\begin{itemize}
    \item \textbf{Mean Reciprocal Rank (MRR)}: The average of the reciprocal ranks of the ground-truth entities.
    \item \textbf{Hits@k}: The proportion of test queries for which the ground-truth entity appears in the top-$k$ predictions. We report Hits@1, Hits@3, and Hits@10.
\end{itemize}

Prior work suggests that the time-aware filtered setting \cite{sun-etal-2021-timetraveler,han2021explainable} is more appropriate than the standard filtered setting \cite{DBLP:conf/ijcai/HeZL0ZZ21} or the raw setting \cite{ijcai2022p284}, because it filters only quadruples that occur at the same timestamp as the query when computing ranks. Therefore, we adopt the time-aware filtered setting for evaluation in all experiments.

\subsubsection{Baseline Methods}
% 比較方法介紹
% We compare our method with several conventional baseline methods, including interpolation and extrapolation TKG reasoning methods. For interpolation methods, we include TTransE \cite{TTtransE}, TA-DistMult \cite{TA-DistMult}, DE-SimplE \cite{Goel_Kazemi_Brubaker_Poupart_2020} and TNTComplEx \cite{TNTComplEx}. For extrapolation methods, we include xERTE \cite{han2021explainable}, TLogic \cite{Tlogic}, CyGNet \cite{CyGNet}, RE-NET \cite{jin-etal-2020-recurrent}, and RE-GCN \cite{RE_GCN}.

We compare our method with several conventional baselines, including xERTE \cite{han2021explainable}, TLogic \cite{Tlogic}, CyGNet \cite{CyGNet}, RE-NET \cite{jin-etal-2020-recurrent}, and RE-GCN \cite{RE_GCN}.

We also compare with prior multi-hop reasoning methods. To make the comparison focused and fair, we keep model capacity comparable and emphasize differences in training procedures rather than architectural choices; the compared agents mainly use LSTM and MLP components. Specifically, we include TITer~\cite{sun-etal-2021-timetraveler}, TAgent~\cite{TAgent}, TPath~\cite{Tpath}, and our ablated variant trained without RAPTOR pretraining (denoted as Pure-RL). To align with our setting, we additionally report a TITer variant with the maximum action size set to 150 (denoted as TITer-150). Aside from these two in-house baselines, the remaining results are taken from previous work~\cite{li-etal-2022-hismatch,Dream}.

\subsection{Experimental Results}
% 實驗結果表格

% 和interpolation的方法比較起來path-based的方法有明顯的優勢，不論在哪個dataset上path-based的方法都具有更好的表現，顯示出path-based的方法對於未見過的timestamp具有更好的處理能力。此外和其他extrapolation的方法比較起來，一般path-based的方法包含RL Only可以達到相當的表現，而RAPTOR進一步提升了結果path-based方法的效果，在ICEWS14和ICEWS05-15資料集上outperform extrapolation methods.雖然在ICEWS18資料集上沒有超越RE-GCN，但RAPTOR仍然達到次佳的效果，顯示出RAPTOR在不同資料集上皆具有穩定的表現。 在比較path-based的方法時，單純使用RL only的方法雖然在ICEWS14表現優於path-based method中表現最好的TITer但在ICEWS05-15和ICEWS18上卻略遜於TITer，相較於RL only的方法，RAPTOR + RL不僅僅在ICEWS14資料集上勝過TITer方法，在ICEWS05-15和ICEWS18亦優於TITer方法。為了更公平的比較，我們亦與相同candidate action space大小TITer的結果進行比較，在action space大小相同(150)的情況下RAPTOR + RL仍然優於TITer，顯示出RAPTOR能有效提升path-based方法的效果。

% Compared to interpolation methods, multi-hop reasoning methods demonstrate a clear advantage, exhibiting superior performance across all datasets. This indicates that multi-hop reasoning methods possess a better capability to predict events at unseen timestamps.

The results are summarized in Table~\ref{tab:main_results}. RL-RAPTOR consistently outperforms Pure-RL across all datasets, highlighting the benefit of reachability pretraining. Compared with prior multi-hop reasoning methods, RL-RAPTOR achieves higher scores on most metrics, suggesting that RAPTOR helps the agent learn better multi-hop reasoning strategies during RL. Among the extrapolation baselines, RL-RAPTOR achieves the best performance on ICEWS14 and ICEWS05-15 on most metrics, while ranking second on ICEWS18 behind RE-GCN. Although Hits@10 is not the best on any of the three datasets, RAPTOR still demonstrates strong and stable improvements over competitive baselines overall.

% Among multi-hop methods, TITer is the strongest prior baseline. RL Only underperforms TITer on ICEWS05-15 and ICEWS18, whereas RL (RAPTOR) surpasses TITer across all datasets. With a matched action space size (\textbf{TITer-150}), RL (RAPTOR) still performs better, underscoring the effectiveness of RAPTOR.

\subsubsection{Training Efficiency Analysis}
% 訓練效率分析、reward曲線比較、validation accuracy收斂速度比較

\begin{figure*}[h]
    \centering
    \includegraphics[width=1.0\linewidth]{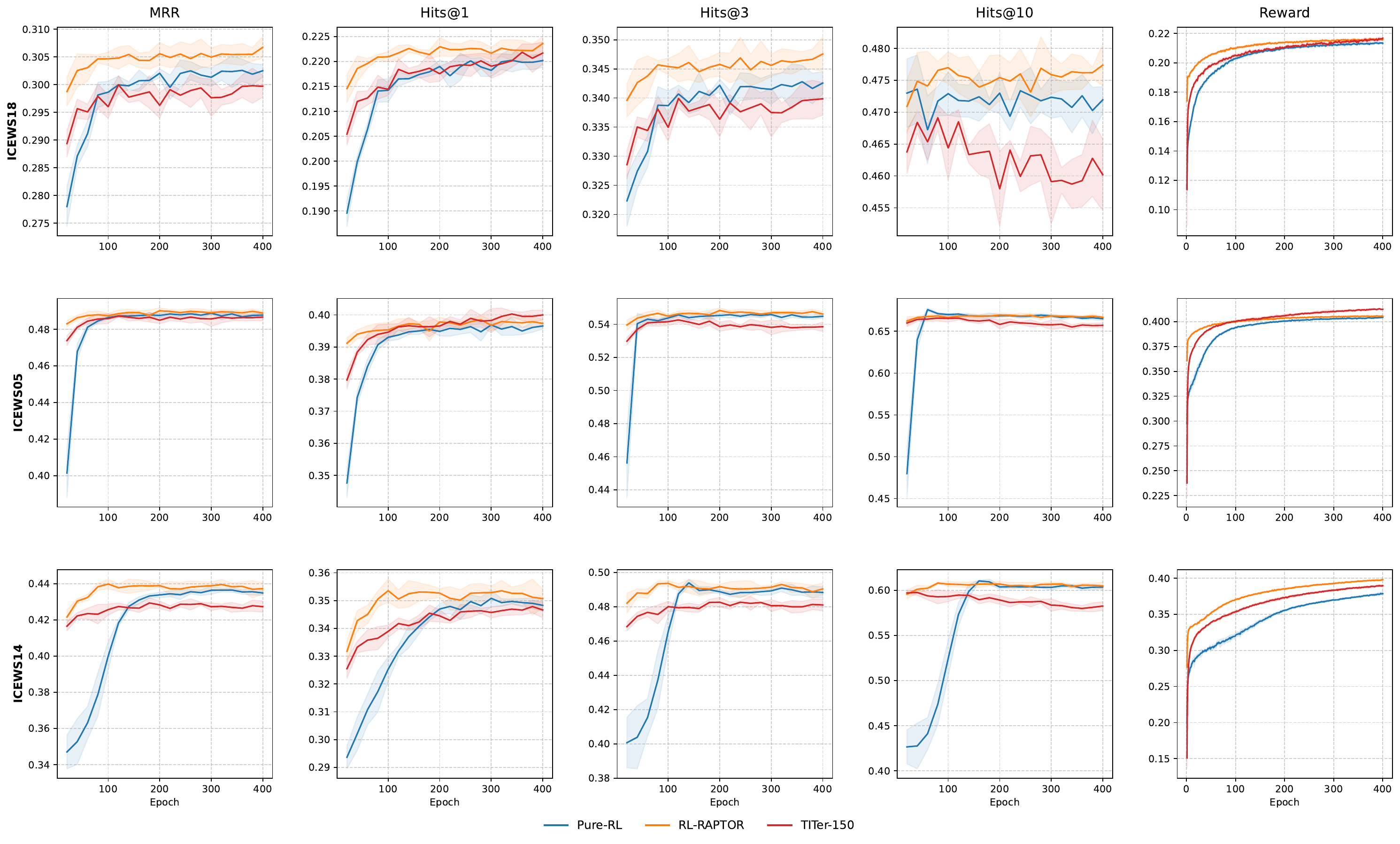}
    \caption{Training efficiency analysis: reward curves during RL training phases on ICEWS14, ICEWS05-15, and ICEWS18 datasets. The shaded areas represent the standard deviation across five random seeds.}
    \label{fig:training_efficiency}
\end{figure*}

We compare training efficiency using validation MRR, Hits@k, and the RL reward curve against Pure-RL and TITer-150. Figure~\ref{fig:training_efficiency} shows that RL-RAPTOR converges earlier and achieves better performance on most metrics. RL-RAPTOR remains best overall and the advantage is more pronounced on harder datasets such as ICEWS18, particularly in MRR.

For Hits@k, RL-RAPTOR is stronger on Hits@1 and Hits@3 across most datasets and attains better score early while staying stable. TITer-150 surpasses RL-RAPTOR in Hits@1 on ICEWS05-15 late in training but declines in Hits@3 and Hits@10 as reward increases, suggesting a shift toward a few high-probability paths. By contrast, RL-RAPTOR maintains stronger Hits@3 and Hits@10 later in training, indicating a more robust action-selection policy. Overall, RAPTOR pretraining improves both training efficiency and final performance, and reaches an early-stage MRR that exceeds the final performance of other baselines.

\subsection{Ablation Studies}
% 單純SL和RL的比較、不同的reachable labeling設定比較

% 在這個section中，我們在ICEWS14資料集上進行消融實驗分析RAPTOR預訓練的epoch數量對最終結果的影響，除了SL epoch以外，我們沿用experiments section中的RL fine-tuning設定在ICEWS14資料集上進行實驗。我們將SL的訓練epoch數分為10, 20, 30, 40, 50五種設定，實驗結果如圖~\ref{fig:ablation_sl_epochs}所示。從Reward曲線的變化，可以很明顯看出隨著pretraining epoch數的增加，模型在RL fine-tuning階段的reward曲線有明顯的提升，且在訓練初期就能獲得較高的reward。此外，在validation MRR曲線中也可以觀察到在RL-finetuning初期，pretraining epoch高的模型明顯能夠更快速的取得較高performance，尤其pretraining epoch為50的模型，在第60個epoch就已經超越不進行pretraining模型的所有validation MRR。此外也可以看出預訓練epoch較高的模型(30, 40, 50 epoch)MRR峰值較高，反之未經預訓練或預訓練epoch較少的10epoch與20epoch峰值明顯較低，但是MRR峰值在30 epoch後的提升並不明顯，顯示出隨著預訓練 Epoch 的增加，RL訓練效率與最佳表現皆有所提昇。即使只訓練10個epoch，模型在RL fine-tuning初期的reward和validation MRR就已經有明顯的提升，這證實了我們透過RAPTOR將reachability inductive bias引入模型中能有效引導 Agent 優化其推理路徑，提升RL stage的訓練效率。

\begin{figure}[h]
    \centering
    \includegraphics[width=1.0\linewidth]{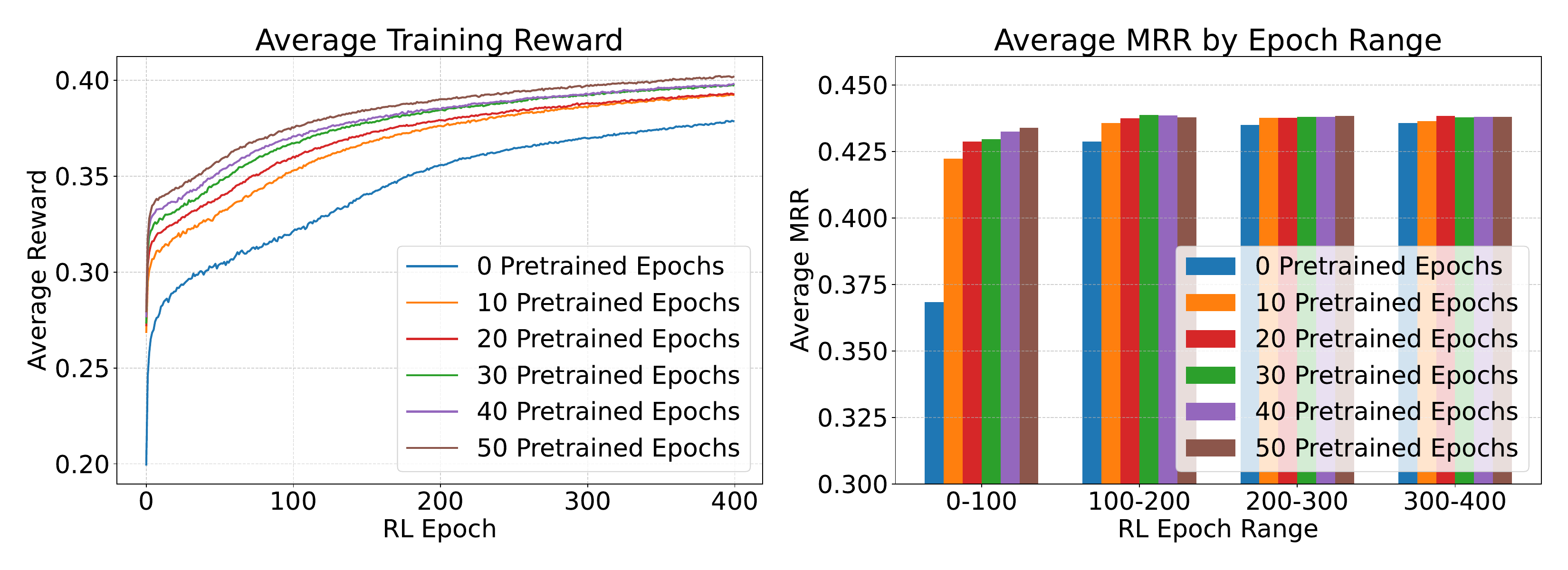}
    \caption{Ablation study on the number of self-supervised pretraining epochs. The left figure shows the training reward and the right figure shows the validation MRR curve during RL fine-tuning.}
    \label{fig:ablation_sl_epochs}
\end{figure}

To assess RAPTOR pretraining, we conduct an ablation study on ICEWS14 and vary the number of pretraining epochs (10, 20, 30, 40, 50). All other hyperparameters in the subsequent RL fine-tuning stage are kept the same as in Appendix~\ref{sec:appendix_impl_details}. Figure~\ref{fig:ablation_sl_epochs} reports the average training reward curve and validation MRR bar chart.

\subsubsection{Analysis of training efficiency and performance}

The reward curves show that longer pretraining benefits the RL stage: models pretrained for more epochs start at higher reward levels, improving sample efficiency. Without pretraining, reward growth is substantially slower. For example, with only 10 epochs of RAPTOR pretraining, the average reward exceeds $0.325$ after 50 epochs (60 total epochs including pretraining), whereas Pure-RL needs more than 100 epochs to reach the same level. These results suggest that RAPTOR pretraining steers the agent toward more promising actions and accelerates learning during RL.

For validation MRR, using more pretraining epochs leads to higher MRR in the early stage of RL training. Although this gap gradually narrows as RL training proceeds, even 10 pretraining epochs already yield a clear improvement over no pretraining, and increasing the number of pretraining epochs further strengthens this advantage. Overall, these results suggest that RAPTOR provides a better initialization, improving both training efficiency and final performance.

\section{Conclusion}

We propose RAPTOR, a self-supervised pretraining method for multi-hop TKG reasoning. By injecting reachability signals, RAPTOR helps the model filter infeasible actions before RL fine-tuning. We also introduce a reachable entity labeling algorithm to identify reachable entities with timestamps and hop distances. Experiments on three benchmarks show improved training efficiency and final performance over conventional baselines. We hope RAPTOR encourages further advances in self-supervised TKG reasoning.

\section{Limitations}
We acknowledge several limitations of this study. First, RAPTOR is currently developed for RL-based multi-hop reasoning frameworks, and its applicability to other paradigms remains underexplored. Second, the reachable entity labeling stage introduces extra preprocessing cost, which may become non-trivial for larger-scale TKGs. Third, due to the limited availability of fully open-source RL-based multi-hop TKG baselines, we use a common and relatively simple backbone to ensure fair and reproducible comparisons. We identify two important and feasible directions for future work: improving the efficiency of the reachable labeling stage, and evaluating RAPTOR with stronger model backbones and more advanced RL training strategies.


\bibliography{custom,my_anthology}

\appendix

% \section{Example Appendix}
% \label{sec:appendix}

% This is an appendix.

\section{Dataset Statistics}
\label{sec:appendix_dataset_stats}

Our datasets are aligned with \citet{li-etal-2022-hismatch} and \citet{Dream}; the dataset statistics are shown in Table~\ref{tab:dataset_statistics}.

\begin{table*}[h]
    \centering
    \caption{Dataset statistics.}
    \label{tab:dataset_statistics}
    \begin{adjustbox}{max width=\textwidth}
    \begin{tabular}{lrrrrrr}
        \toprule
        Dataset & \#Entities & \#Relations & \#Train & \#Valid & \#Test & Time granularity\\
        \midrule
        ICEWS14 & 7,128 & 230 & 74,845 & 8,514 & 7,371 & 1 Day \\
        ICEWS05-15 & 10,094 & 251 & 368,868 & 46,302 & 46,159 & 1 Day \\
        ICEWS18 & 23,033 & 256 & 373,018 & 45,995 & 49,545 & 1 Day \\
        \bottomrule
    \end{tabular}
    \end{adjustbox}
\end{table*}

\section{Implementation Details}
\label{sec:appendix_impl_details}

In our experiments, the entity and relation embedding dimensions are set to $128$, with a temporal component of size $d_t=48$ in the relation embedding. The step embedding dimension is $d_{\text{step}}=32$, and the maximum path length is $K=3$ for all datasets. Before self-supervised pretraining, we perform reachable-entity labeling with hop limit $K$ and retain the top-$N=200$ reachable entities.

We use RAPTOR to pretrain the agent for 40 epochs and then fine-tune with RL for 400 epochs. During RL, the initial entropy coefficient is $\beta_0=0.01$ with exponential decay factor $\zeta=0.9$, and the discount factor is $\gamma=0.95$. We use a batch size of 512 in both phases and optimize with Adam at a learning rate of $10^{-3}$. At each time step, we keep the top-150 candidate actions sorted by timestamp during both pretraining and RL.

For checkpoint selection, we evaluate every 20 epochs using beam search with beam size 100 and select the checkpoint with the best validation performance. All experiments use five random seeds, and we report the mean across seeds.

\section{Actor-Critic Optimization Details}
\label{sec:appendix_actor_critic}

We train the policy (actor) and value (critic) networks using an actor-critic algorithm.
The actor parameterizes a stochastic policy $\pi_\theta(a \mid s)$, while the critic estimates the state value $V_\phi(s)$.
The value function serves as a baseline to estimate the expected return from $s$ and reduces the variance of policy gradients.

For time step $k \in \{0, \ldots, K-1\}$ in an episode, we compute the discounted return at time step $k$ as
\begin{equation}
    g_k = \gamma^{(K-1)-k} R,
    \label{eq:return}
\end{equation}
where $\gamma \in (0,1]$ is the discount factor.
The advantage is defined by
\begin{equation}
    Adv_k = g_k - V_\phi(s).
    \label{eq:advantage}
\end{equation}

We update the actor by maximizing the expected advantage:
\begin{equation}
    \mathcal{L}_{\mathrm{actor}}(\theta)
    = \mathbb{E}\left[\sum_{k=0}^{K-1} \log \pi_\theta(a_{k} \mid s)\, Adv_k \right].
    \label{eq:actor}
\end{equation}
The critic is trained by minimizing the squared error between the predicted value and return:
\begin{equation}
    \mathcal{L}_{\mathrm{critic}}(\phi)
    = \mathbb{E}\left[\sum_{k=0}^{K-1} \left(V_\phi(s) - g_k\right)^2\right].
    \label{eq:critic}
\end{equation}

In addition, we incorporate an entropy regularization term to encourage exploration:
\begin{equation}
    \mathcal{L}_{\mathrm{entropy}}(\theta)
    = \mathbb{E}\left[\sum_{k=0}^{K-1} H\!\left(\pi_\theta(\cdot \mid s)\right)\right],
    \label{eq:entropy}
\end{equation}
where $H(\cdot)$ denotes the entropy of the action distribution.

The overall objective is
\begin{equation}
    \max_{\theta}\mathcal{L}_{\mathrm{actor}}(\theta) + \beta_n\, \mathcal{L}_{\mathrm{entropy}}(\theta),
    \;
    \min_{\phi}\mathcal{L}_{\mathrm{critic}}(\phi),
    \label{eq:overall}
\end{equation}
where $\beta_n$ is an entropy coefficient at training epoch $n$.
We apply an exponential decay schedule $\beta_n = \beta_0 \zeta^{\,n}$ with decay factor $\zeta \in (0,1)$.
We optimize $\theta$ and $\phi$ jointly using stochastic gradient descent.

\section{Disclosure of AI Use}

The authors used GitHub Copilot as a coding assistant during implementation. ChatGPT and Gemini were used during manuscript preparation for writing support, including improving clarity, grammar, and phrasing. These tools were not used to generate the core scientific contributions, make research decisions, or determine the experimental results. All methods, experiments, analyses, and conclusions were developed, verified, and approved by the authors, who take full responsibility for the content of the paper.

\end{document}